\documentclass[10pt,twocolumn,letterpaper]{article}
\pdfoutput=1

\usepackage{iccv}
\usepackage{times}
\usepackage{epsfig}
\usepackage{amsmath}
\usepackage{amssymb}
\usepackage{graphicx}
\DeclareMathOperator*{\argmax}{arg\,max}

\usepackage[breaklinks=true,bookmarks=false]{hyperref}

\iccvfinalcopy 

\def\iccvPaperID{****} 

\ificcvfinal\fi

\begin{document}

\title{Weakly-Supervised HOI Detection from Interaction Labels Only and Language/Vision-Language Priors}

\author{Mesut Erhan Unal\\
Department of Computer Science\\
University of Pittsburgh\\
{\tt\small meu6@pitt.edu}
\and
Adriana Kovashka\\
Department of Computer Science\\
University of Pittsburgh\\
{\tt\small kovashka@cs.pitt.edu}
}

\maketitle
\ificcvfinal\thispagestyle{empty}\fi

\begin{abstract}
Human-object interaction (HOI) detection aims to extract interacting human-object pairs and their interaction categories from a given natural image. Even though the labeling effort required for building HOI detection datasets is inherently more extensive than for many other computer vision tasks, weakly-supervised directions in this area have not been sufficiently explored due to the difficulty of learning human-object interactions with weak supervision, rooted in the combinatorial nature of interactions over the object and predicate space. In this paper, we tackle HOI detection with the weakest supervision setting in the literature, using only image-level interaction labels, with the help of a pretrained vision-language model (VLM) and a large language model (LLM). We first propose an approach to prune non-interacting human and object proposals to increase the quality of positive pairs within the bag, exploiting the grounding capability of the vision-language model. Second, we use a large language model to query which interactions are possible between a human and a given object category, in order to force the model not to put emphasis on unlikely interactions. Lastly, we use an auxiliary weakly-supervised preposition prediction task to make our model explicitly reason about space. Extensive experiments and ablations show that all of our contributions increase HOI detection performance.
\end{abstract}




\section{Introduction}
\label{sec:intro}

\begin{figure}[!t]
    \includegraphics[width=1.0\linewidth]{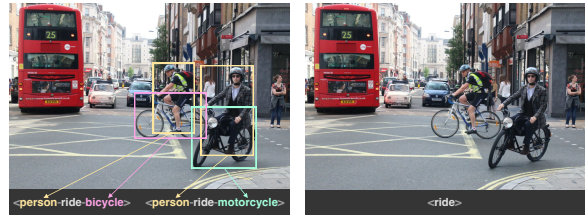}
    \label{fig:title_figure}
    \vspace{-4mm}
    \caption{\textbf{Top-left:} Existing HOI detection methods need costly annotations which contain bounding boxes for interacting human-object pairs as well as their interaction categories. \textbf{Top-right:} Our method relies on image-level interaction labels, without any information on where, between whom and how many times those interactions occur. \textbf{Bottom:} During training, our approach utilizes image captions to prune non-interacting human/object proposals with the help of a vision-language model. Remaining human and object proposals will be paired for classification and a large language model will verify if predicted interactions are plausible. Best viewed in color with zoom.}
\end{figure}

Human-object interaction (HOI) detection is formally defined as correctly localizing interacting human-object pairs and classifying their interaction in a given natural image.
The problem has been formulated in different ways, either end-to-end, or more commonly 
as a two-step procedure wherein all human and object instances get detected first, then interacting human-object pairs are identified.
Regardless of its formulation, researchers rely on strong supervision to tackle HOI detection.
This strong supervision is in the form of bounding box annotations for interacting human-object pairs as well as semantic labels for their interactions, which are costly to acquire and cognitively demanding for annotators as they require one to fully understand the image content\footnote{\url{http://www-personal.umich.edu/~ywchao/hico/hoi-det-ui/demo_20171121.html}}.
Despite the excessive cost of gathering annotations for HOI detection, weakly-supervised directions to relax this strong supervision need have not been fully explored, 
due to the combinatorial complexity of object interactions over object and predicate space.

In this paper, we tackle weakly-supervised HOI detection using the weakest supervision in the literature, namely image-level interaction labels (e.g. ride).
This supervision level is less costly and more natural to acquire than ones required in existing efforts, as annotators would be required to answer a simple question: ``What are the individuals doing in this picture?''.  
To make learning possible, our approach utilizes free-form captions paired with images to weakly-supervise an auxiliary task and to prune non-interacting humans and objects.
We query a large language model (LLM) to eliminate unlikely human-object interactions (e.g. riding toothbrush).
To increase the spatial reasoning capability of our model, we further formulate an auxiliary preposition prediction task.
In this task, our model learns to assign one of the predefined prepositions 
to each human-object pair during training via weak supervision.
Having free-form captions in hand also gives us the ability of extracting image-level interaction labels using a language parser, and hence further relax the level of supervision.
Our code will be released upon publication.
To summarize, our main contributions are as follows:

\begin{itemize}
\item We formulate a weakly-supervised HOI detection setting where supervision comes from image-level interaction labels (e.g. ride, eat). This weak supervision has not previously been used in the literature.
\item We utilize free-form captions paired with images to exploit the implicit grounding capability of a vision language model (VLM) in order to prune non-interacting human and object proposals.
\item We make use of an large language model (LLM) to verify if a given $<$interaction, object$>$ pair is plausible.
\item To further increase our model's spatial reasoning capability, we formulate a weakly-supervised preposition prediction task. 
\item For the first time in the literature, we train an HOI detection model using image-caption pairs which are abundant on the web. 

\end{itemize}
\section{Related Work}
\subsection{Human-object interaction detection}
The problem of detecting interactions between humans and objects was originally introduced in \cite{gupta2015visual} and has drawn immense attention in the computer vision community since then.
Most of the research efforts on this topic \cite{gao2018ican,li2019transferable,ulutan2020vsgnet,kim2020detecting,li2020hoi,zhang2021spatially} use a two-stage solution in which human/object locations are extracted along with their semantic labels by an off-the-shelf object detector first, and an interaction classification model is learnt on pairwise human-object features.
Apart from human/object appearances, there exist models that make use of contextual features \cite{gao2018ican,ulutan2020vsgnet}, spatial layouts \cite{li2019transferable,ulutan2020vsgnet,zhang2021spatially} and human pose estimations \cite{li2019transferable}.
Inspired by one-stage object detection efforts, researchers lately try to formulate end-to-end HOI detection approaches where human/instance localization and interaction classification are performed in parallel \cite{liao2020ppdm,kim2020uniondet,kim2021hotr,kim2022mstr}.
These methods are analogous to CNN-based (e.g. YOLO\cite{redmon2016you}) and Transformer-based (e.g. DETR\cite{carion2020end}) end-to-end object detectors.
PPDM\cite{liao2020ppdm} takes a step forward and drops the need for heuristically created ``anchors'', formulating HOI detection as a point matching problem between human and object locations.

Regardless of being one-stage or two-stage, these methods rely on strong supervision which is costly to acquire.
This supervision is in the form of quadruplets that contain interacting human-object locations, object category and interaction category.
Even though HOI detection is extremely costly to supervise, there exists a lack of weakly-supervised efforts in the literature. 
Among the existing weakly-supervised methods, MX-HOI \cite{kumaraswamy2021detecting} proposes a momentum-independent learning framework where they utilize both weak and strong supervision.
Additionally, AlignFormer \cite{kilickaya2021human} formulates an alignment layer in transformer framework, that generates pseudo-aligned human-object pairs from weak annotations, conditioning on geometric and visual priors.
Both of these methods utilize image-level $<$interaction, object$>$ annotations (e.g. \{eat-banana\}) as weak supervision as opposed to the much weaker supervision we use in our work, namely image-level interaction labels (e.g. \{eat\}).

\subsection{Using cues from vision-language models}
Following vision-language models' breakthrough, researchers have explored their usage in aiding diverse computer vision tasks.
For example, one of the most popular VLMs, namely CLIP \cite{radford2021learning}, has been researched extensively in the context of image generation \cite{patashnik2021styleclip}, cross-modal retrieval \cite{bai2021connecting,dzabraev2021mdmmt}, image classification \cite{bai2021connecting}, object detection \cite{gu2022openvocabulary}, HOI detection \cite{dong2022category,liao2022genvltk} and image captioning \cite{changpinyo2021conceptual}, thanks to its robust image-text joint space learned on a massive dataset.

Even though \cite{dong2022category,liao2022genvltk} also utilize CLIP in the context of HOI detection, how CLIP is employed within our approach is quite different.
\cite{dong2022category} uses CLIP's text encoder to initialize context-aware HOI queries within a fully-supervised Transformer-based HOI detector.
\cite{liao2022genvltk} utilize CLIP as a teacher within their model and distill knowledge for both visual and textual understanding of interactions.
The most similar work to ours in terms of how CLIP is employed is ProposalCLIP \cite{shi2022proposalclip}, where authors prune low-quality object proposals produced by a static algorithm (e.g. EdgeBoxes \cite{zitnick2014edge}).
Their method runs cropped proposal regions along with produced captions for object categories (i.e. $\{\text{``a photo of a } c_i^{(obj)}\text{''}\}_{i=1}^{|C^{(obj)}|}$) through CLIP and removes proposals based on the alignment entropy over caption set.
In our work, on the other hand, we need to quantify if a given proposal is a part of an interaction or not.
Unlike running CLIP on large number of proposals, we run whole image through CLIP once and create grounding maps to calculate an interaction score on each proposal.

\begin{figure*}[!t]
    \centering
    \includegraphics[width=\linewidth]{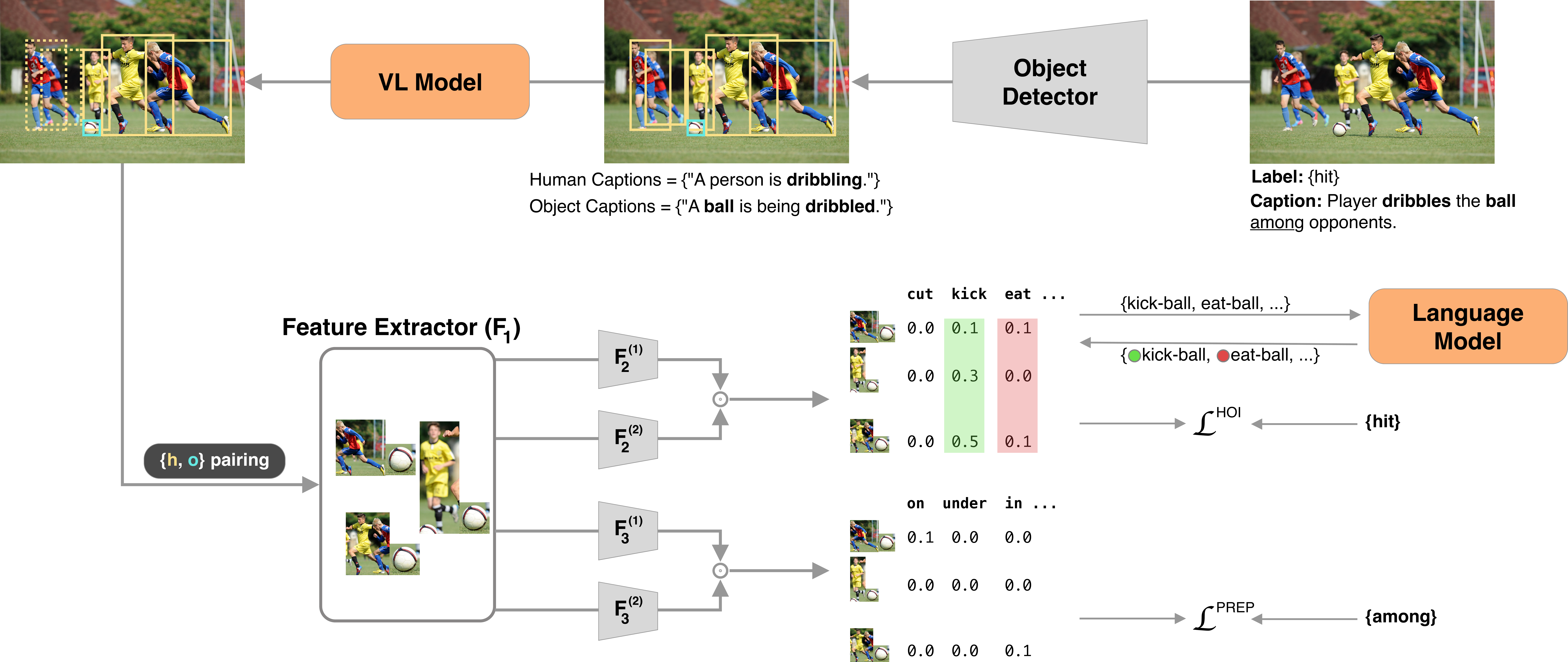}
    \label{fig:paper_figure}
    \vspace{-4mm}
    \caption{\textbf{Overview of our method during training.} Retrieving human and object proposals from an object detector, our method first prunes non-interacting human/object proposals with the help of a vision-language model, calculating an interaction score for each proposal. Next, we pair remaining human-object proposals and run those pairs through a two-stream feed-forward neural net ($F_2$) that operates on $F_1$'s output space. Finally, image-level predictions are calculated by summing $F_2$'s output over region pairs. We query a large-language model to restrict our model's output space only to meaningful interactions. In order to improve our model's spatial reasoning capability, we formulate a weakly-supervised preposition prediction task wherein supervision comes from preposition extracted from captions. During inference, we drop proposal pruning and preposition prediction modules, requiring only an image to detect HOI instances.}
\end{figure*}

\section{Method}
\subsection{Formulating HOI detection with weak supervision}
\label{subsec:wshoi}

Assume an object detector that outputs a set of human and object predictions for a given image, $\mathcal{H} = \{h_i\}_{i=1}^N$ and $\mathcal{O} = \{o_j\}_{j=1}^M$ respectively.
Each of these predictions is in the form of $\{x^{(1)}, y^{(1)}, x^{(2)}, y^{(2)}, c^o, s^o\}$ where $x^{(1)},y^{(1)}$ and $x^{(2)},y^{(2)}$ denote the top-left and bottom-right corner coordinates of the proposal bounding box respectively, $c^o$ is the semantic category assigned by the object detector (``person'' for each proposal in $\mathcal{H}$) and $s^o \in [0, 1]$ is the confidence score.

Given (1) the above set of human and object proposals, and (2) provided image-level interaction labels (e.g. {ride}), our goal is to learn an HOI detection model, $F(\cdot)$, that can map each human-object pair $\{h, o\} \in \mathcal{H} \times \mathcal{O}$ to an interaction class $c^v$ that belongs to a predefined set of classes $C^v = \{c^{v, (k)}\}_{k=1}^K$ and a confidence score $s^v \in [0, 1]$, yielding $\mathcal{H} \times \mathcal{O} \xrightarrow{F(\cdot)} \{c^v_{i,j}, s^v_{i,j}\}_{i=1, j=1}^{N, M}$.
Here the $v$ superscript denotes ``verb'' (interaction), while $k$ denotes the specific verb/interaction class.
Please note that this definition of weakly-supervised HOI detection is slightly different than the one given in \S Sec. \ref{sec:intro} as we offload localization and semantic labeling of humans and objects to an object detector as in every two-stage HOI detection work.
We can further rewrite $F$ as composition of two separate functions, $F_1$ and $F_2$, where $F_1$ is responsible for extracting pairwise features and modeling interactions while $F_2$ performs classification, yielding $F = F_2 \circ F_1$.

In the fully-supervised case, learning is facilitated by giving the model access to the correct HOI targets $Y = \{\text{bbox}^{\text{human}}_i, \text{bbox}^{\text{object}}_i, c^o_i, c^v_i\}_{i=1}^L$ which contain ground-truth human and object locations as bounding box coordinates, as well as the semantic categories for the object and the interaction.
When one has ground-truth targets at hand, an HOI detection model can be trained to increase the likelihood of $\{h, o\}$ pairs that spatially and semantically overlap with a HOI target to have the same interaction class as that target.
In our case, however, the model can only access a set of ground-truth interaction classes for a given image, without even knowing if a certain interaction happens once or multiple times in the image (see Figure \ref{fig:title_figure}).

Inspired by existing weakly-supervised object detection (WSOD) literature, we formulate weakly-supervised HOI detection as a multiple instance learning (MIL) problem.
In this formulation, each image is considered as a bag of human-object pairs (i.e. $\mathcal{H} \times \mathcal{O}$).
If a bag (i.e. image) is labeled positive for a certain interaction, it has to contain at least one $\{h, o\}$ pair of that interaction.
Similar to WSDDN \cite{bilen2016weakly}, we split the final classification layer, $F_2$, into a two-stream head (i.e. $F_2^{(1)}$ and $F_2^{(2)}$) where one models ``what is the most probable interaction class for a given human-object pair?'' (i.e. $P(C^v \,| \, \{h, o\},\, F_2^{(1)})$) while the other models ``what is the most probable human-object pair for a given interaction class?'' (i.e. $P(\{h_i, o_j\}_{i=1, j=1}^{N, M} \,| \, c^v,\, F_2^{(2)})$).
Assuming we get $d$-dimensional feature for each pair through $F_1$ as row vectors in a feature matrix $\mathbf{Z}$, such as $F_1(\mathcal{H} \times \mathcal{O}) = \mathbf{Z} \in \mathbb{R}^{NM \times d}$, the aforementioned probabilities can be calculated by mapping $\mathbf{Z}$ to a $|C^v|$-dimensional space first and then applying softmax on different dimensions.
Hence, we can formulate $F_2^{(1)}$ as a mapping $\mathbb{R}^d \rightarrow \mathbb{R}^{|C^v|}$ followed by a row-wise softmax on $\mathbf{Z}$ while
$F_2^{(2)}$ can be formulated as a mapping $\mathbb{R}^d \rightarrow \mathbb{R}^{|C^v|}$ followed by a column-wise softmax on $\mathbf{Z}$.
Then, we can write $F_2$'s output $\mathbf{Z}^{\text{HOI}}$ as:
\begin{equation}
    \begin{split}
        \mathbf{Z}^{\text{HOI}} &= F_2^{(1)}(\mathbf{Z}) \odot F_2^{(2)}(\mathbf{Z}) \in \mathbb{R}^{NM \times |C^v|} \qquad \text{where} \\
        &F_2^{(1)}(\mathbf{Z}) = \sigma^{\rightarrow}(\mathbf{Z}*W_{F_2^{(1)}}) \\
        &F_2^{(2)}(\mathbf{Z}) = \sigma^{\downarrow}(\mathbf{Z}*W_{F_2^{(2)}})
    \end{split}
\end{equation}
where $\odot$ represents Hadamard product, $N$ is the number of human proposals, $M$ is the number of object proposals, $|C^v|$ is the number of interaction classes, $W_{F_2^{(1)}}, W_{F_2^{(2)}} \in \mathbb{R}^{d \times |C^v|}$ are weight matrices, and $\sigma^{\rightarrow}$ and $\sigma^{\downarrow}$ are row-wise and column-wise softmax operations, respectively.

Finally, we can formulate our learning objective as minimizing $|C^v|$ binary classification losses, one for each interaction class $c^v \in C^v$:

\begin{equation}
    \label{eq:wshoi}
    \begin{split}
        \mathcal{L}^{\text{HOI}}(\hat{Y}^v, Y^v) &= \frac{1}{|C^v|} \sum_{k=1}^{|C^v|} \ell (\hat{y}^{v, (k)}, y^{v, (k)}) \\
        \hat{Y}^v &= \sum^{NM} \mathbf{Z}^{\text{HOI}}
    \end{split}
\end{equation}
where $Y^v$ is the binary image-level interaction labels and $y^{v, (k)} = 1$ iff $c^{v, (k)}$ is apparent in the image, $0$ otherwise.
Let $\mathbf{Z}^{\text{HOI}}_{[i \times j]}$ denote the $|C^v|$-dimensional row vector in $\mathbf{Z}^{\text{HOI}}$ that corresponds to the pair $\{h_i, o_j\}$. Then one can obtain the pair's interaction class $c^v_{i,j}$ and confidence score $s^v_{i,j}$ during inference as:

\begin{equation}
    \label{eq:inference}
    \begin{split}
        c^v_{i,j} &= c^{v, (\argmax_{k}\mathbf{Z}^{\text{HOI}, (k)}_{[i \times j]})} \\
        s^v_{i,j} &= \max(\mathbf{Z}^{\text{HOI}}_{[i \times j]}) \cdot s_{h_i}^o \cdot s_{o_j}^o
    \end{split}
\end{equation}
where $s_{h_i}^o$ and $s_{o_j}^o$ are confidence scores assigned to $h_i$ and $o_j$ by the object detector, respectively.

\subsection{Extracting interaction labels from captions}
\label{subsec:extract}
Our weakly-supervised HOI detector learning procedure requires image-level interaction labels for supervision.
However, one can utilize captions to extract those annotations to further relax the level of annotation required.
In this work, we demonstrate how one can train an HOI detector on a dataset scraped from the web and contains noisy captions, using a simple technique.

We start with extracting nouns and verbs from captions using a POS tagger \cite{Honnibal_spaCy_Industrial-strength_Natural_2020}.
Consider we have a set of predefined interaction categories $C^v$, and verb and noun sets for a particular image $\mathcal{V}=\{v_i\}_{i=1}^A$ and $\mathcal{N}=\{n_i\}_{i=1}^B$, respectively.
For each image where ``person'' $\in \mathcal{N}$, we construct its label as $Y^v = \{v \, | \, v \in \mathcal{V} \, \text{and} \, v \in C^v\}$.
We use a synonym list to match ``person'', as given in \cite{lu2018neural}.

\subsection{Pruning non-interacting proposals}
\label{subsec:pruning}

Learning an HOI detector using only image-level interaction labels is inherently a difficult task. A model needs to learn how to identify interacting human-object pairs among a large candidate pool and classify their interactions correctly.
Without bounding box supervision, the model is left by itself to learn how an interacting human or object should look like, and what combination of those maps to a certain interaction class.
For instance, consider learning the interaction ``kick'' with object ``ball''.
We can expect that most of the images containing this particular interaction and object would portray a game field where more than one person is apparent.
\textbf{To a weakly-supervised model, each person would be equally likely to be the subject of the ``kick'' interaction.}
One can try to build coarse heuristic rules (e.g. interacting human-object pairs should be close in space) or more fine-grained ones (e.g. human-object pairs for kick interaction should be close in space, but they can be further for another interaction) to reduce the search space but it is impossible to precisely develop rules for every natural interaction.
To this end, we propose to exploit the implicit grounding capability of a vision-language model 
to prune non-interacting human and object proposals.
In this work, we employ CLIP \cite{radford2021learning} and produce visual grounding maps for image-text pairs using \cite{chefer2021generic}.

Consider that we have access to free-form captions for the images in our training data; we do \emph{not} require captions at inference time.
Given an image and its caption, we first extract all verbs $\mathcal{V}=\{v_i\}_{i=1}^A$ and nouns $\mathcal{N}=\{n_i\}_{i=1}^B$ out of the caption using a POS tagger \cite{Honnibal_spaCy_Industrial-strength_Natural_2020}.
We then create human captions as $HC = \{\text{``a person is $v_i$-ing''}\}_{i=1}^A$ and object captions as $OC = \{\text{``a $n_i$ is being $v_j$-ed''}\}_{i=1, j=1}^{B, A}$.
We run the image and created captions through CLIP to produce a grounding map per caption and resize them into original image dimensions via bilinear interpolation.
Finally, grounding maps are min-max normalized to map their values into $[0, 1]$ range.

Retrieving grounding maps $GH$ and $GO$ for human and object captions respectively, one can calculate a grounding score, $g$, for each human and object proposal, $h \in \mathcal{H}, o \in \mathcal{O}$.
Intuitively, $g$ should measure how likely a certain proposal is to engage in an interaction.
We calculate $g$ for each proposal as follows:

\begin{equation}
    \small
    \begin{split}
        g_{h} &= \frac{1}{(x^{(2)}_{h} - x^{(1)}_{h})(y^{(2)}_{h} - y^{(1)}_{h})} \frac{1}{|GH|} \sum_{k=1}^{|GH|} \sum_{i=x^{(1)}_{h}, j=y^{(1)}_{h}}^{x^{(2)}_{h}, y^{(2)}_{h}} GH_{i, j}^{(k)} \\
        g_{o} &= \frac{1}{(x^{(2)}_{o} - x^{(1)}_{o})(y^{(2)}_{o} - y^{(1)}_{o})} \frac{1}{|GO|} \sum_{k=1}^{|GO|} \sum_{i=x^{(1)}_{o}, j=y^{(1)}_{o}}^{x^{(2)}_{o}, y^{(2)}_{o}} GO_{i, j}^{(k)}
    \end{split}
\end{equation}

The above equations simply calculate the average grounding score that falls into each human/object proposal region using the corresponding grounding maps.
Finally, the interaction score for a human proposal $h_i$ or an object proposal $o_j$ is calculated as the multiplication of its grounding score $g$ and confidence score given by the object detector $s^o$:  

\begin{equation}
    I_{h_i} = g_{h_i} \cdot s_{h_i}^o \, , \, I_{o_j} = g_{o_j} \cdot s_{o_j}^o
\end{equation}

The reason behind multiplying $g$ and $s^o$ for interaction score calculation is quite simple.
In our experiments, we have seen that the generated grounding maps usually focus on the most distinct parts of the interacting human/object, which would result in proposals covering only those distinct areas to get the highest interaction scores if only $g$ was used.

Lastly, we sort human/object proposals in descending order of their interaction scores $I_h$/$I_o$ and keep top $50\%$ as is while assigning a special ``background'' class to others.
These ``background'' proposals still get paired with human proposals within the model and will serve as negatives.

\subsection{Suppressing implausible interactions}
\label{subsec:supressing}

Previous work \cite{gupta2019no,zhang2021spatially} has shown that it can be beneficial to restrict a model's output space only to meaningful interactions, conditioning on some type of lookup table in which plausible interactions are encoded.
While \cite{gupta2019no} proposed to learn these conditions within the model optimizing an indicator function over possible interactions given human and object proposals, \cite{zhang2021spatially} compute them directly on data, iterating over ground-truth HOI targets.
There also exist works that learn such conditions by modeling interactions as phrases (e.g. ``person eat banana'') 
in a textual \cite{lu2016visual} or multi-modal space \cite{Ye_2021_CVPR}.
Unlike these methods, ours does not require subject-predicate-object annotations nor multimodal training.


In this work, we propose to use a large language model (LLM) to query which interactions are plausible for a given object category.
Our hypothesis is that these models would have learnt natural co-occurrences throughout their training on massive text, and this information would also be applicable to the visual domain.
We consider two natural approaches for how an LLM can be used for this purpose: (1) inputting ``A person is [MASK] $c^o$'' caption to the model (where $c^o$ denotes a particular object category) and calculating a probability distribution over possible interaction categories $C^v = \{c^{v, (k)}\}_{k=1}^K$ at the masked-language modeling (MLM) head to obtain the [MASK] token, ignoring the rest of the vocabulary, and (2) plugging ``What a person do with $c^o$?'' as a question and interaction classes $C^v = \{c^{v, (k)}\}_{k=1}^K$ as an answer set, then retrieving the language model's output distribution over $C^v$ at the multiple choice question answering (MCQA) head.
After obtaining a probability distribution over interaction classes given an object category i.e. $P(C^v \, | \, c^o$), we create a binary lookup table for each object category, wherein interaction categories are encoded as plausible (if their probability is larger than average) or otherwise implausible.

\begin{equation}
    \begin{split}
        \Phi_{c^o} &= \{\phi(c^o, c^{v, (k)})\}_{k=1}^{|C^v|} \\
        \phi(c^o, c^{v, (k)}) &= \left\{
            \begin{array}{ll}
              1 & P(c^{v, (k)} \, | \, c^o) > \frac{1}{|C^v|} \sum^{|C^v|}P(c^v \, | \, c^o) \\
              0 & \text{otherwise} \\
            \end{array} 
        \right.
    \end{split}
\end{equation}

Lastly, we double the confidence score of a human-object pair $\{h_i, o_j\}$ if its predicted label is plausible given the object category of $o_j$:

\begin{equation}
    s^{v'}_{i,j} = s^v_{i, j} \cdot (1 + \phi(c_{o_j}^o, c_{i, j}^v))
\end{equation}

We use RoBERTa \cite{liu2019roberta} to query if a given $<$interaction, object$>$ pair is plausible.
We build $\Phi_{c^o}$ for each dataset before training, instead of querying LLM constantly.

\subsection{Formulating weakly-supervised preposition prediction}
\label{subsec:wsppp}
Prior work \cite{kant2020spatially} demonstrates that encoding pairwise spatial relations as discrete labels (e.g. inside of, contains) within a model improves performance on tasks that require explicit spatial understanding, such as TextVQA \cite{singh2019towards}.
Inspired from but different from their work, we formulate a preposition prediction task in which the model is forced to learn a mapping from pairwise features to discrete spatial labels in weakly-supervised manner, in the unique context of human-object interaction.

Similar to our weakly-supervised HOI detection formulation given in \S Sec. \ref{subsec:wshoi}, we employ a two-stream head, $F_3$, that operates on $F_1$'s output space.
Assuming our pre-defined preposition set is $C^p = \{c^{p, (k)}\}_{k=1}^{K}$ and we get a $d$-dimensional feature for each human-object pair through $F_1$ as row vectors in feature matrix $\textbf{Z}$, such as $F_1(\mathcal{H} \times \mathcal{O}) = \mathbf{Z} \in \mathbb{R}^{NM \times d}$, we formulate $F_3^{(1)}$ as a mapping $\mathbb{R}^d \rightarrow \mathbb{R}^{|C^p|}$ followed by a row-wise softmax on $\mathbf{Z}$ while $F_3^{(2)}$ is formulated as a mapping $\mathbb{R}^d \rightarrow \mathbb{R}^{|C^p|}$ followed by a column-wise softmax on $\mathbf{Z}$.
Then, we can write $F_3$'s output $\mathbf{Z}^{\text{PREP}}$ as:
\begin{equation}
    \begin{split}
        \mathbf{Z}^{\text{PREP}} &= F_3^{(1)}(\mathbf{Z}) \odot F_3^{(2)}(\mathbf{Z}) \in \mathbb{R}^{NM \times |C^p|} \qquad \text{where} \\
        &F_3^{(1)}(\mathbf{Z}) = \sigma^{\rightarrow}(\mathbf{Z}*W_{F_3^{(1)}}) \\
        &F_3^{(2)}(\mathbf{Z}) = \sigma^{\downarrow}(\mathbf{Z}*W_{F_3^{(2)}})
    \end{split}
\end{equation}
where $\odot$ represents Hadamard product, $N$ is the number of human proposals, $M$ is the number of object proposals, $|C^p|$ is the number of preposition classes, $W_{F_3^{(1)}}, W_{F_3^{(2)}} \in \mathbb{R}^{d \times |C^p|}$ are weight matrices, and $\sigma^{\rightarrow}$ and $\sigma^{\downarrow}$ are row-wise and column-wise softmax operations, respectively.

Finally, we formulate our learning objective as minimizing $|C^p|$ binary classification losses, one for each preposition class $c^p \in C^p$:

\begin{equation}
    \label{eq:wsppp}
    \begin{split}
        \mathcal{L}^{\text{PREP}}(\hat{Y}^p, Y^p) &= \frac{1}{|C^p|} \sum_{k=1}^{|C^p|} \ell(\hat{y}^{p, (k)}, y^{p, (k)}) \\
        \hat{Y}^p &= \sum^{NM} \mathbf{Z}^{\text{PREP}}
    \end{split}
\end{equation}
where $Y^p$ is the binary image-level preposition labels and $y^{p, (k)} = 1$ iff $c^{p, (k)}$ is apparent in the image, $0$ otherwise.
Adding a new task in the model, our overall training objective now becomes minimizing both $\mathcal{L}^{\text{HOI}}$ (Eq. \ref{eq:wshoi}) and $\mathcal{L}^{\text{PREP}}$:

\begin{equation}
    \label{eq:finalloss}
    \mathcal{L} = \mathcal{L}^{\text{HOI}} + \lambda \mathcal{L}^{\text{PREP}}
\end{equation}

Since none of the datasets we use in our experiments comes with such preposition annotations, we utilize captions to extract them.
Specifically, we run captions through a scene graph parser (e.g. Stanford Scene Graph Parser \cite{schuster2015generating}) to extract $<$subject, predicate, object$>$ triplets.
We filter out triplets whose subject is not ``person'' or predicate is not in $C^p$, which we curated by hand collecting $32$ most common prepositions.
We use the same synonym list for ``person'', mentioned in \S Sec. \ref{subsec:extract}.
After the filtering process, the unique predicates from the remaining triplets are used as image-level preposition labels for their corresponding images.

\section{Experiments}

\subsection{Setup}

\textbf{Datasets and metrics.} We use the well-established HOI detection benchmark datasets, HICO-DET \cite{chao2018learning} and V-COCO \cite{gupta2015visual}, in our experiments.
HICO-DET contains $37,633$ training and $9,546$ test images with bounding box annotations for interacting human-object pairs and their interaction labels.
There are $80$ object (same as in MS COCO \cite{lin2014microsoft}) and $117$ interaction categories in HICO-DET with $600$ unique $<$interaction, object$>$ pairs.
As HICO-DET instances do not come with a paired caption, we use a state-of-the-art image captioning model, OFA \cite{wang2022ofa}, to generate one for each image in the training split.
On the other hand, V-COCO is a relatively smaller dataset with $5,400$ images in trainval and $4,946$ images in test split.
There are $80$ object and $26$ interaction categories.
As V-COCO is a subset of MS COCO, each image is paired with $5$ captions. 
We use standard metrics for each dataset which are Agent AP and Role AP for V-COCO, and Full mAP for HICO-DET.
Please note that HICO-DET's Full mAP is analogous to V-COCO's Role AP, which requires predicted human and object bounding boxes to have at least $0.5$ IoU with corresponding HOI target, and predicted interaction category should be the same as the target interaction label.
V-COCO's Agent AP, on the other hand, requires correct localization of humans (i.e. $\text{IoU} > 0.5$) engaging in a particular interaction.

Furthermore, for the first time in the literature we learn an HOI detection model on a small subset of Conceptual Captions, which consists of roughly $18,000$ image-caption pairs, without any HOI-related annotation.
We extract image-level interaction labels from captions as explained in \S Sec. \ref{subsec:extract}.
We use the V-COCO test split to evaluate models trained on this new dataset.

\textbf{Baseline and training procedure.} Please note that our proposed approach as a whole is applicable to any existing two-stage HOI detection method.
In our experiments, we use SCG \cite{zhang2021spatially} as our baseline and implement our main contributions on top of it.
We choose SCG because it is one of the best performing fully-supervised two-stage HOI detector with a publicly-available implementation.
Unless noted otherwise, we use the same hyperparameter settings as \cite{zhang2021spatially}.
We use Faster R-CNN\cite{ren2015faster} with ResNet50-FPN\cite{He_2016_CVPR} pretrained on MS COCO to generate detections.
We train all models on $4 \times$ NVIDIA Quadro RTX 5000 GPUs with an initial learning rate of $1e-4$ and batch size of $16$ ($4$ images per GPU).
On V-COCO and HICO-DET, all models are trained for $8$ epochs, reducing learning rate to $1e-5$ after $6$\textsuperscript{th} epoch.
On the other hand, we train models on Conceptual Captions subset for $5$ epochs, without applying any decay strategy on learning rate.
For weakly-supervised HOI detection task, we use binary adaptation of focal loss \cite{lin2017focal} ($\ell$ in Eq. \ref{eq:wshoi}) following the baseline, and binary cross entropy loss for weakly-supervised preposition prediction task ($\ell$ in Eq. \ref{eq:wsppp}).
We set the weight for weakly-supervised preposition prediction task as $0.1$ ($\lambda$ in Eq. \ref{eq:finalloss}).
Interested readers may consult the original paper for additional details on model implementation and training procedure.

\begin{table}[!t]
    \small
    \begin{tabular}{ll p{1.50cm} p{1.50cm}}
        Method & Sup. & Backbone & \multicolumn{1}{c}{Role AP} \\ \hline \hline
        \multicolumn{1}{l|}{iCAN \cite{gao2018ican}} & \multicolumn{1}{l}{Full} & \multicolumn{1}{l|}{RN50} & \multicolumn{1}{c}{52.04} \\
        \multicolumn{1}{l|}{VSGNet \cite{ulutan2020vsgnet}} & \multicolumn{1}{l}{Full} & \multicolumn{1}{l|}{RN152} & \multicolumn{1}{c}{57.00} \\
        \multicolumn{1}{l|}{SCG \cite{zhang2021spatially}} & \multicolumn{1}{l}{Full} & \multicolumn{1}{l|}{RN50 FPN} & \multicolumn{1}{c}{58.02} \\
        \multicolumn{1}{l|}{IDN \cite{li2020hoi}} & \multicolumn{1}{l}{Full} & \multicolumn{1}{l|}{RN50} & \multicolumn{1}{c}{60.30} \\
        \multicolumn{1}{l|}{HOTR \cite{kim2021hotr}} & \multicolumn{1}{l}{Full} & \multicolumn{1}{l|}{RN50+Transformer} & \multicolumn{1}{c}{64.40} \\
        \multicolumn{1}{l|}{MSTR \cite{kim2022mstr}} & \multicolumn{1}{l}{Full} & \multicolumn{1}{l|}{RN50+Transformer} & \multicolumn{1}{c}{\textbf{65.20}} \\ \hline
        \multicolumn{1}{l|}{MX-HOI \cite{kumaraswamy2021detecting}} & \multicolumn{1}{l}{Weak+} & \multicolumn{1}{l|}{RN101} & \multicolumn{1}{c}{-} \\
        \multicolumn{1}{l|}{AlignFormer \cite{kilickaya2021human}} & \multicolumn{1}{l}{Weak+} & \multicolumn{1}{l|}{RN50} & \multicolumn{1}{c}{\textbf{14.15}} \\ \hline
        \multicolumn{1}{l|}{Baseline \cite{zhang2021spatially} (\S\ref{subsec:wshoi})} & \multicolumn{1}{l}{Weak} & \multicolumn{1}{l|}{RN50 FPN} & \multicolumn{1}{c}{20.05} \\
        \multicolumn{1}{l|}{Ours} & \multicolumn{1}{l}{Weak} & \multicolumn{1}{l|}{RN50 FPN} & \multicolumn{1}{c}{\textbf{29.59}} \\ \hline
        \multicolumn{1}{l|}{Ours-CC} & \multicolumn{1}{l}{Weak-} & \multicolumn{1}{l|}{RN50 FPN} & \multicolumn{1}{c}{\textbf{17.71}} \\ \hline
    \end{tabular}
    \caption{V-COCO test Role AP performance among methods trained on V-COCO trainval split (except OursCC). Ours outperforms AlignFormer by a large margin (absolute $15.54\%$), even though its supervision comes from image-level $<$interaction, object$>$ labels (\textbf{Weak+}) rather than image-level $<$interaction$>$ only labels (\textbf{Weak}) we use. It also greatly improves (absolute $9.54\%$) over Baseline, which is close to SOTA when trained fully-supervised, verifying effectiveness of our contributions. Trained on a dataset scraped from the web, extracting image-level $<$interaction$>$ only labels from captions (\textbf{Weak-}), our method (OursCC) still outperforms AlignFormer by absolute $3.56\%$. RN denotes ResNet. MX-HOI did not report V-COCO results. \textbf{Bolding} shows the best method within each supervision level.}
    \label{table:vcoco}
\end{table}

\subsection{Comparison with the SOTA}
In this subsection, we compare our model against the state-of-the-art HOI detection efforts.
We also include fully-supervised approaches to inform readers on the performance gap between fully- and weakly-supervised HOI detection literature, and to show that our baseline SCG \cite{zhang2021spatially} is comparable with SOTA when trained with strong supervision.
We would like to stress that there are not many weakly-supervised work in the HOI detection literature and existing approaches (e.g. MX-HOI \cite{kumaraswamy2021detecting} and AlignFormer \cite{kilickaya2021human}) use image-level $<$interaction, object$>$ annotations as weak supervision.
Careful readers would have already noticed that this specific definition of ``weak supervision'' is considerably stronger than is in our formulation as it reduces the search space on object proposals to match with a specific interaction category in a given image. 
Consider a natural image that portrays a person riding a bike while another riding a motorcycle in an urban setting.
For that particular image, image-level $<$interaction, object$>$ annotation will yield to \{ride-bike, ride-motorcycle\} while image-level interaction annotation (our supervision) will be \{ride\} only.
If $<$interaction, object$>$ annotation are exploited, object proposals that can be matched with ``ride'' will be narrowed down to bike and motorcycle.
In our case, however, the object space is excessively larger, including other objects that may be apparent in the image (e.g. bus, car, etc.).

In Table \ref{table:vcoco}, we compare HOI detection performance on V-COCO test split among the models trained on the V-COCO trainval set (except Ours-CC).
Our method improves absolute $9.54\%$ over weakly-supervised variant of SCG, which we build our contributions upon, and absolute $15.54\%$ over AlignFormer, which uses stronger supervision in the form of image-level $<$interaction, object$>$ labels.
Our method trained on the Conceptual Captions subset (Ours-CC) also surpasses AlignFormer and achieves a comparable performance to weakly-supervised SCG, even though we extract image-level interaction labels from captions to supervise its learning (\S Sec. \ref{subsec:extract}) and use MS COCO trained object detector to produce human/object proposals.

\begin{table}[!t]
    \small
    \begin{tabular}{ll p{1.50cm} p{1.50cm}}
        Method & Sup. & Backbone & \multicolumn{1}{c}{mAP} \\ \hline \hline
        \multicolumn{1}{l|}{iCAN \cite{gao2018ican}} & \multicolumn{1}{l}{Full} & \multicolumn{1}{l|}{RN50} & \multicolumn{1}{c}{14.84} \\
        \multicolumn{1}{l|}{VSGNet \cite{ulutan2020vsgnet}} & \multicolumn{1}{l}{Full} & \multicolumn{1}{l|}{RN152} & \multicolumn{1}{c}{19.80} \\
        \multicolumn{1}{l|}{SCG \cite{zhang2021spatially}} & \multicolumn{1}{l}{Full} & \multicolumn{1}{l|}{RN50 FPN} & \multicolumn{1}{c}{21.85} \\
        \multicolumn{1}{l|}{IDN \cite{li2020hoi}} & \multicolumn{1}{l}{Full} & \multicolumn{1}{l|}{RN50} & \multicolumn{1}{c}{23.36} \\
        \multicolumn{1}{l|}{HOTR \cite{kim2021hotr}} & \multicolumn{1}{l}{Full} & \multicolumn{1}{l|}{RN50+Transformer} & \multicolumn{1}{c}{23.46} \\
        \multicolumn{1}{l|}{MSTR \cite{kim2022mstr} \dag} & \multicolumn{1}{l}{Full} & \multicolumn{1}{l|}{RN50+Transformer} & \multicolumn{1}{c}{\textbf{31.17}} \\ \hline
        \multicolumn{1}{l|}{MX-HOI \cite{kumaraswamy2021detecting} \dag} & \multicolumn{1}{l}{Weak+} & \multicolumn{1}{l|}{RN101} & \multicolumn{1}{c}{16.14} \\
        \multicolumn{1}{l|}{AlignFormer \cite{kilickaya2021human} \dag} & \multicolumn{1}{l}{Weak+} & \multicolumn{1}{l|}{RN50} & \multicolumn{1}{c}{\textbf{19.26}} \\ \hline
        \multicolumn{1}{l|}{Baseline \cite{zhang2021spatially} (\S\ref{subsec:wshoi})} & \multicolumn{1}{l}{Weak} & \multicolumn{1}{l|}{RN50 FPN} & \multicolumn{1}{c}{7.05} \\
        \multicolumn{1}{l|}{Ours} & \multicolumn{1}{l}{Weak} & \multicolumn{1}{l|}{RN50 FPN} & \multicolumn{1}{c}{\textbf{8.38}} \\ \hline
    \end{tabular}
    \caption{Full mAP (Default) comparison on HICO-DET. Unsurprisingly, AlignFormer and MX-HOI benefit from having much stronger supervision, namely image-level $<$interaction, object$>$ labels (\textbf{Weak+}), when combinatorial complexity over interaction space is increased moving from V-COCO to HICO-DET. However, Ours still improves over Baseline, which uses the same image-level $<$interaction$>$ only labels (\textbf{Weak}) verifying effectiveness of our contributions. RN denotes ResNet. \textbf{\dag} denotes using an object detector fine-tuned on HICO-DET.}
    \label{table:hicodet}
\end{table}

\begin{figure*}[!t]
    \includegraphics[width=\linewidth]{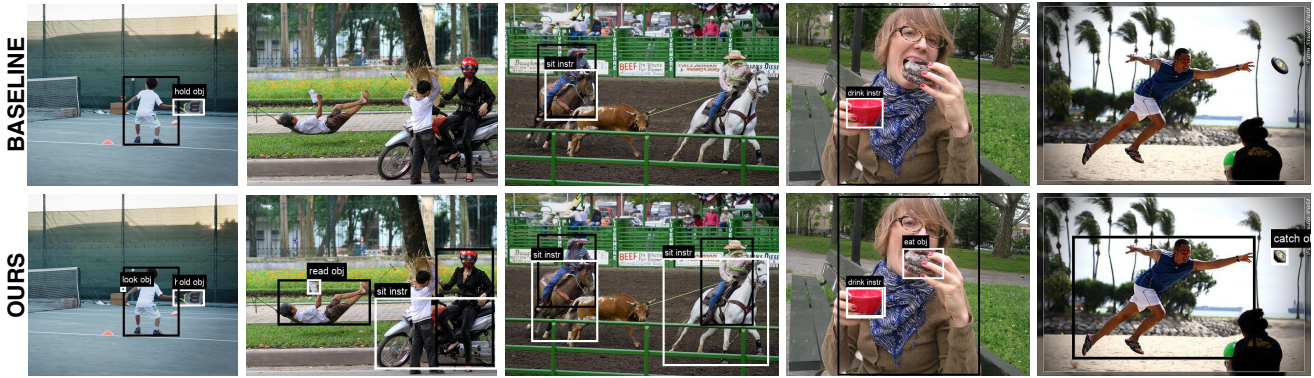}
    \label{fig:qualitative_figure}
    \caption{\textbf{Qualitative examples} sampled from V-COCO test split. Black and white boxes show interacting humans and objects, respectively. Our method successfully detects more interactions than Baseline, especially when the same human is subject to more than one interactions. Moreover, the 3\textsuperscript{rd} example shows that it selects the better object proposal for ``sit'' interaction (horse). Interaction label explanations can be found in original V-COCO paper \cite{gupta2015visual}, Table 1.}
\end{figure*}

Similarly in Table \ref{table:hicodet}, we compare HOI detection performance on the HICO-DET test split among the models trained on the HICO-DET training set.
Results show that both weakly-supervised SCG and Ours fail to sustain their improvement over AlignFormer, due to the increased number of interaction categories over V-COCO (26 vs 117).
Unsurprisingly, AlignFormer was not affected heavily by increased combinatorial complexity over the $<$interaction, object$>$ joint space thanks to its stronger supervision than ours.
It is also worth noting that both AlignFormer and MX-HOI use an object detector fine-tuned on HICO-DET (denoted by $\dag$) while we do not.
Regardless of its unsustained performance on HICO-DET, our method still improves over the baseline weakly-supervised SCG by absolute $1.33\%$ (relative $18.87\%$), which has been trained with the same level of supervision as Ours.

\subsection{Ablation study}
To demonstrate effectiveness of our contributions, we incrementally ablate them over the baseline weakly-supervised SCG on V-COCO, HICO-DET and Conceptual Captions. The results are shown in Tables \ref{table:ablationvcoco}, \ref{table:ablationhico} and \ref{table:ablationconcap}.
While all of our contributions clearly improve the performance over the baseline, results also show that caption-dependent parts of our method (\S Sec. \ref{subsec:pruning} \& \S Sec. \ref{subsec:wsppp}) are not affected heavily from the caption source.
Independent of whether captions are collected in a controlled setting (V-COCO), scraped from the web (Conceptual Captions) or generated by a captioning model (HICO-DET), our model can utilize them to boost model performance.

\begin{table}[t]
    \small
    \begin{tabular}{ll p{1.50cm} p{1.50cm}}
        Method & Sup. & \multicolumn{1}{c}{Agent AP ($\Delta$)} & \multicolumn{1}{c}{Role AP ($\Delta$)} \\ \hline \hline
        \multicolumn{1}{l|}{Baseline \cite{zhang2021spatially} (\S\ref{subsec:wshoi})} & \multicolumn{1}{c|}{Weak} & \multicolumn{1}{c}{32.41} & \multicolumn{1}{c}{20.05} \\ \hline
        \multicolumn{1}{l|}{+Pruning (\S\ref{subsec:pruning})} & \multicolumn{1}{c|}{Weak} & \multicolumn{1}{c}{33.88 \color{green}(+1.47)} & \multicolumn{1}{c}{21.80 \color{green}(+1.75)}\\
        \multicolumn{1}{l|}{+Suppressing (\S\ref{subsec:supressing})} & \multicolumn{1}{c|}{Weak} & \multicolumn{1}{c}{37.04 \color{green}(+4.63)} & \multicolumn{1}{c}{28.28 \color{green}(+8.23)} \\
        \multicolumn{1}{l|}{+Preposition (\S\ref{subsec:wsppp})} & \multicolumn{1}{c|}{Weak} & \multicolumn{1}{c}{40.53 \color{green}(+8.12)} & \multicolumn{1}{c}{29.59 \color{green}(+9.54)} \\ \hline
    \end{tabular}
    \caption{Incremental ablations on V-COCO. $\Delta$ denotes performance difference over Baseline. All three of our contributions help improving HOI detection performance.}
    \label{table:ablationvcoco}
\end{table}

\begin{table}[t]
    \centering
    \small
    \begin{tabular}{lll}
        Method & Sup. & \multicolumn{1}{c}{mAP ($\Delta$)} \\ \hline \hline
        \multicolumn{1}{l|}{Baseline \cite{zhang2021spatially} (\S\ref{subsec:wshoi})} & \multicolumn{1}{c|}{Weak} & \multicolumn{1}{c}{7.05} \\ \hline
        \multicolumn{1}{l|}{+Pruning (\S\ref{subsec:pruning})} & \multicolumn{1}{c|}{Weak} & \multicolumn{1}{c}{7.55 \color{green}(+0.50)} \\
        \multicolumn{1}{l|}{+Suppressing (\S\ref{subsec:supressing})} & \multicolumn{1}{c|}{Weak} & \multicolumn{1}{c}{7.81 \color{green}(+0.76)} \\
        \multicolumn{1}{l|}{+Preposition (\S\ref{subsec:wsppp})} & \multicolumn{1}{c|}{Weak} & \multicolumn{1}{c}{8.38 \color{green}(+1.33)} \\ \hline
    \end{tabular}
    \caption{Incremental ablations on HICO-DET. $\Delta$ denotes performance difference over Baseline. All three of our contributions help improving HOI detection performance.}
    \label{table:ablationhico}
\end{table}

\begin{table}[t]
    \small
    \begin{tabular}{ll p{1.50cm} p{1.50cm}}
        Method & Sup. & \multicolumn{1}{c}{Agent AP ($\Delta$)} & \multicolumn{1}{c}{Role AP ($\Delta$)} \\ \hline \hline
        \multicolumn{1}{l|}{Baseline \cite{zhang2021spatially} (\S\ref{subsec:wshoi})} & \multicolumn{1}{c|}{Weak} & \multicolumn{1}{c}{17.71} & \multicolumn{1}{c}{14.33} \\ \hline
        \multicolumn{1}{l|}{+Pruning (\S\ref{subsec:pruning})} & \multicolumn{1}{c|}{Weak} & \multicolumn{1}{c}{19.44 \color{green}(+1.73)} & \multicolumn{1}{c}{15.95 \color{green}(+1.62)}\\
        \multicolumn{1}{l|}{+Suppressing (\S\ref{subsec:supressing})} & \multicolumn{1}{c|}{Weak} & \multicolumn{1}{c}{20.00 \color{green}(+2.29)} & \multicolumn{1}{c}{18.23 \color{green}(+3.90)} \\
        \multicolumn{1}{l|}{+Preposition (\S\ref{subsec:wsppp})} & \multicolumn{1}{c|}{Weak} & \multicolumn{1}{c}{20.75 \color{green}(+3.04)} & \multicolumn{1}{c}{17.71 \color{green}(+3.38)} \\ \hline
    \end{tabular}
    \caption{Incremental ablations on Conceptual Captions. $\Delta$ denotes performance difference over Baseline. While all three contributions help improve performance over Baseline,  the preposition prediction task slightly decreases Role AP when added on top of implausible interaction suppression (but still boosts Agent AP).}
    \label{table:ablationconcap}
\end{table}

\section{Conclusion}
In this work, we tackle HOI detection problem with the weakest supervision setting in the literature, using image-level interaction labels only (e.g. ``ride'').
We exploit the implicit grounding capability of a vision-language model, 
in order to prune non-interacting human and object proposals.
We restrict our model's output space to natural interactions only, querying a large language model if a given $<$interactions, object$>$ is plausible.
We lastly formulate a weakly-supervised preposition prediction task to improve spatial reasoning capability of our model explicitly.
For the first time in the literature, we learn an HOI detector on image-caption pairs, extracting image-level interaction labels out of captions.

\noindent \textbf{Ethical concerns.} VLMs and LLMs can contain implicit biases inherited from their training data.
Even though their usage within this work's context did not pose any explicit harm during our experimentation, we would like to warn users that their usage in a different context may expose people to potentially unethical content.

{\small
\bibliographystyle{ieee_fullname}
\bibliography{final-camera}
}

\end{document}